%% file: main.tex
\documentclass[journal]{IEEEtran} 
\usepackage{amsmath,amsfonts}
\usepackage{algorithmic}
\usepackage{algorithm}
\usepackage{array}
\usepackage[caption=false,font=normalsize,labelfont=sf,textfont=sf]{subfig}
\usepackage{textcomp}
\usepackage{stfloats}
\usepackage{url}
\usepackage{verbatim}
\usepackage{graphicx}
\usepackage{amsmath}
\usepackage{amssymb}
\usepackage{booktabs}
\usepackage{hhline}
\usepackage{pifont}
\usepackage{multirow}
\usepackage{tabu}
\usepackage[table,xcdraw]{xcolor}
\usepackage{dsfont}
\usepackage{bbm}
\usepackage{cite}
\usepackage{csquotes}
\usepackage[capitalize]{cleveref}
\crefname{section}{Sec.}{Secs.}
\Crefname{section}{Section}{Sections}
\Crefname{table}{Table}{Tables}
\crefname{table}{Tab.}{Tabs.}

\newcommand{\tableautorefname}{Table}

\newcommand{\cmark}{\ding{51}}%
\newcommand{\xmark}{\ding{55}}%

\newcommand{\argmin}[1]{\underset{#1}{\operatorname{arg}\,\operatorname{min}}\;}

\begin{document}

\title{InstaGraM: \textbf{Insta}nce-level \textbf{Gra}ph \textbf{M}odeling
\\for Vectorized HD Map Learning}

\author{Juyeb Shin, Hyeonjun Jeong,~\IEEEmembership{Graduate Student Member,~IEEE},
Francois Rameau, Dongsuk Kum,~\IEEEmembership{Member,~IEEE}

\thanks{
Manuscript received 20 December 2023; revised 5 June 2024 and 6 October 2024; accepted 14 November 2024. This work was supported by Institute of Information and communications Technology Planning and Evaluation (IITP) and the National Research Foundation of Korea (NRF) grant funded by the Korea government (MSIT) (No.2021-0-00951, Development of Cloud based Autonomous Driving AI learning Software and 2022R1A2C2004944). \textit{(Corresponding author: Dongsuk Kum.)}
}
\thanks{
Juyeb Shin is with The Robotics Program, Korea Advanced Institute of Science and Technology, Daejeon 34051, Republic of Korea (e-mail:juyebshin@kaist.ac.kr) 
}
\thanks{
Hyeonjun Jeong and Dongsuk Kum are with the Graduate School of Mobility, Korea Advanced Institute of Science and Technology, Daejeon 34051, Republic of Korea (e-mail:\{hyeonjun.jeong, dskum\}@kaist.ac.kr) 
}
\thanks{
Francois Rameau is with the Computer Science department, State University of New York - Korea, Songdo 21985, Republic of Korea (e-mail: francois.rameau@sunykorea.ac.kr)
}
\thanks{This paper extends our work presented in conference workshop~\cite{shin2023instagram} with further experiments and analysis including quantitative and qualitative comparison with state-of-the-art methods, and ablation studies.}
}

\markboth{IEEE TRANSACTIONS ON INTELLIGENT TRANSPORTATION SYSTEMS,~Accepted}%
{Shin \MakeLowercase{\textit{et al.}}: InstaGraM: Instance-level Graph Modeling for Vectorized HD Map Learning}

\IEEEpubid{0000--0000/00\$00.00~\copyright~2021 IEEE}

\maketitle

\input{Section/0_abs}
\input{Section/1_intro}
\input{Section/2_related}
\input{Section/3_method}
\input{Section/4_exp}
\input{Section/5_conc}

\bibliographystyle{IEEEtran} 
\bibliography{IEEEfull}


\input{Section/6_bio}

\vfill

\end{document}

%% file: Section/0_abs.tex
\begin{abstract}
For scalable autonomous driving, a robust map-based localization system, independent of GPS, is fundamental.
To achieve such map-based localization, online high-definition (HD) map construction plays a significant role in accurate estimation of the pose.
Although recent advancements in online HD map construction have predominantly investigated on vectorized representation due to its effectiveness, they suffer from computational cost and fixed parametric model, which limit scalability.
To alleviate these limitations, we propose a novel HD map learning framework that leverages graph modeling.
This framework is designed to learn the construction of diverse geometric shapes, thereby enhancing the scalability of HD map construction.
Our approach involves representing the map elements as an instance-level graph by decomposing them into vertices and edges to facilitate accurate and efficient end-to-end vectorized HD map learning.
Furthermore, we introduce an association strategy using a Graph Neural Network to efficiently handle the complex geometry of various map elements, while maintaining scalability.
Comprehensive experiments on public open dataset show that our proposed network outperforms state-of-the-art model by $1.6$ mAP.
We further showcase the superior scalability of our approach compared to state-of-the-art methods, achieving a $4.8$ mAP improvement in long range configuration.
Our code is available at {https://github.com/juyebshin/InstaGraM}.
\end{abstract}


%% file: Section/1_intro.tex
\section{Introduction}\label{sec:intro}
\IEEEPARstart{A}{ccurate} and robust vehicle localization is one of the most essential pre-requisites for deploying autonomous driving.
Traditionally, localization system has mostly adopted either GPS and IMU measurements to obtain global position, or signal-level information such as visual features (SIFT~\cite{lowe2004distinctive}, ORB~\cite{rublee2011orb}, etc.) and LiDAR intensities along with SLAM methods.
However, the former system that depends on high cost GPS limits the scalability of autonomous driving, caused by the signal blocking and drift problems, meanwhile the latter system lacks robustness in the dynamic large-scale environments.
Recent localization approaches have shown great alternatives by leveraging semantic and geometric understanding of the scene from onboard sensors~\cite{lu2017monocular,guo2021coarse,zhang2021dt,li2021bsp,wang2021visual,qin2021light,zhang2022bevlocator,jeong2024multi}.
These approaches utilize static road environment for map-matching to achieve robust and scalable localization.

\begin{figure}[t]
\centering
\includegraphics[width=0.9\linewidth]{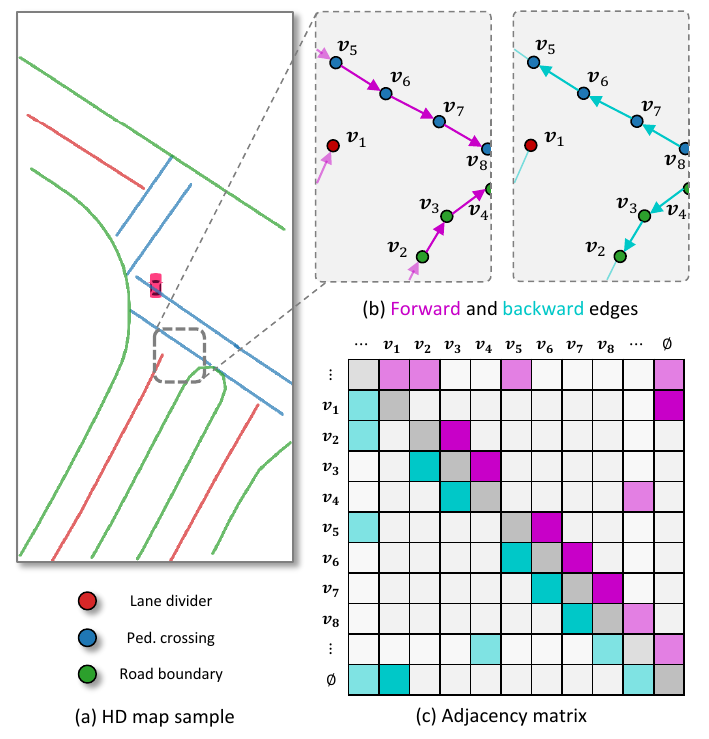}
\vspace{-5pt}
\caption{We model vectorized map elements as a graph, composed of vertices and edges.
(a) is an HD map sample taken from \textit{nuScenes}.
(b) illustrates the vertices of map elements and their bidirectional edges, depicted as forward and backward.
}
\label{fig:1_Graph}
\vspace{-12pt}
\end{figure}

\begin{figure*}[tb]
\centering
\includegraphics[width=0.9\linewidth]{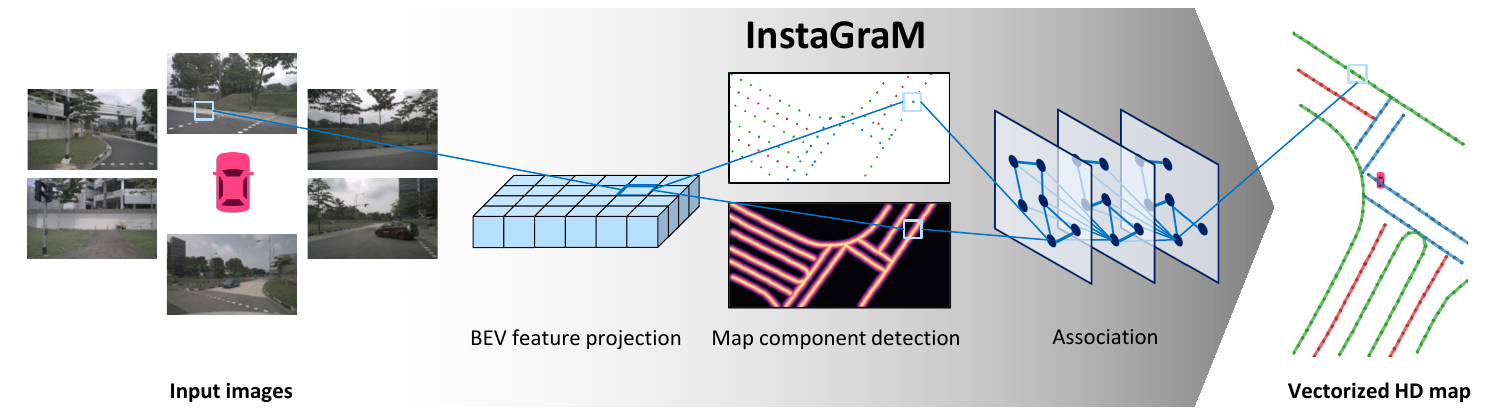}
\vspace{-5pt}
\caption{We propose InstaGraM, a hybrid architecture of CNNs and a GNN for real-time HD map learning in bird's-eye-view representation. Starting from the input surround images and camera parameters, a unified BEV representation is extracted by projecting and fusing image features. InstaGraM extracts vertex locations and implicit edge maps of map elements, and final vectorized HD map elements are generated throughout a GNN.}
\label{fig:2_Pipeline}
\vspace{-12pt}
\end{figure*}

\IEEEpubidadjcol
Understanding static road environment via online HD map construction plays a significant role for localization of a vehicle.
Early works of HD map construction view the task as segmentation problem, which predicts the occupancy of the rasterized world grids~\cite{philion2020lift,pan2020cross,roddick2020predicting,li2022bevformer,peng2022bevsegformer,zhou2022cross,chen2022efficient}.
While these approaches demonstrate their relevance, such representation remains memory intensive and lacks geometric details desirable not only for map-matching localization but also for downstream tasks such as trajectory planning and motion forecasting.
As its counterpart, vectorized representation~\cite{liang2018end,liang2019convolutional,homayounfar2019dagmapper,li2021hdmapnet,liu2022vectormapnet,liao2023maptr} maintains structural information of the map, providing instance and geometric attributes, suitable for autonomous driving.
Relevant work adopts various intermediate outputs to construct vectorized representation of HD map, which requires post-processing with heavy computational cost~\cite{li2021hdmapnet}.
To further bring the vectorized HD map construction end-to-end learnable without any post-processing, recent works utilize direct set prediction using Transformer decoder~\cite{carion2020end}.
Despite their compelling performances, those methods remain computationally demanding due to autoregressive models~\cite{liu2022vectormapnet} or inherently suffer from scalability problem due to fixed parametric geometries by its design~\cite{liao2023maptr}.
To address the aforementioned limitations, we propose end-to-end vectorized HD map learning, namely Instance-level Graph Modeling (InstaGraM).
InstaGraM views the vectorized HD map as a graph construction problem.
A graph with the vertices and their connections best suits vectorized HD map representation that it can preserve geometric and instance-level information of the map elements with no computational cost, as illustrated in Fig.~\ref{fig:1_Graph}.
On top of novel instance-level graph modeling of vectorized HD map, we propose an HD map construction model that learns to construct the graph representation of the map, composed of three stages, as depicted in Fig.~\ref{fig:2_Pipeline}.
First, we extract the feature maps from each image --- captured by the camera rig --- and aggregate them into a single top-down feature map using a 2D-to-BEV transformation~\cite{pan2020cross,li2021hdmapnet}.
Then, given this top-down feature map, we detect the road elements' vertex points and edge maps via CNNs. 
Finally, the vertices' positions and their respective local edge maps are passed to a GNN to predict their instance-level connections as an adjacency matrix trained in a supervised manner.
Throughout graph learning scheme with graph modeling and GNN, our model is capable of capturing various shapes and length of map instances, enhancing scalable prediction.
Our contributions can be summarized as:
\begin{itemize}
\item{We propose a novel graph modeling for vectorized HD map elements that models geometric, semantic and instance-level information as graph representations.}
\item{On top of the proposed graph modeling, we present an end-to-end vectorized HD map learning network with GNN that enables accurate and fast construction while maintaining scalability.}
\end{itemize}

%% file: Section/2_related.tex
\section{Related Work}
\label{sec:related_works}
\subsection{Online HD Map Construction}
Online HD map construction task has recently been widely explored and developed.
This task falls into two main categories by their representations: a rasterized representation that approaches map construction as a segmentation problem, and a vectorized representation that produces structured model of the map elements.
HD map construction model utilizes encoder-decoder architecture where encoder extracts a unified BEV feature map from sensor inputs and then decoder produces map output from the BEV feature map.
\subsubsection{Rasterized map segmentation}
This task mainly focuses on the camera input modality for its rich semantic information with additional fusion with LiDAR for 3D information.
Specifically for the camera modality, main challenges of the rasterized map segmentation problem is the ill-posed view transformation from perspective view (PV) images to BEV feature map.
Thanks to the recent advances of previous works, rasterized map segmentation has gained relevant development.
When only BEV space is considered, such representation can easily be obtained via Inverse Perspective Mapping (IPM)~\cite{mallot1991inverse} of the perspective view feature map~\cite{ReiherLampe2020Cam2BEV,hou2020multiview} -- assuming known camera intrinsics and extrinsics.
This IPM warping is valid under the planar assumption but is violated for semantic clues of any object above the ground level (\textit{e.g.}, cars and pedestrians) potentially leading to severe perspective distortion in the resulting map elements.
To avoid such stretching effect,~\cite{zhu2018generative,mani2020monolayout} utilize Generative Adversarial Network (GAN) which learns cross-view transformation between PV and BEV.

Alternatively, to recover full structural information in 3D space, some approaches utilize depth information~\cite{philion2020lift,li2023bevdepth,liu2023bevfusion,kim2023crn}.
Lift strategy in~\cite{philion2020lift} predicts categorical depth distribution along the pixel rays and combines with 2D CNN feature to obtain voxel-like representation which is then collapsed along z-axis to BEV space.
BEV pooling~\cite{liu2023bevfusion} boosts the computation of aforementioned splat operation using interval reduction and fast grid association.
BEVDepth~\cite{li2023bevdepth} further exploits supervision of depth distribution for reliable depth acquisition.
CRN~\cite{kim2023crn} adopts sparse radar point clouds on top of BEVDepth for more spatially accurate BEV feature.
Homography or depth-aware view transform strategies have the advantage of being intuitive, interpretable and offering good transferability to various camera setups.
Despite these advantages, geometric warping solutions face multiple limitations: they rely on strong prior, may suffer from perspective distortions and require successive stages.
To circumvent these limitations, another solution consists in using neural networks to learn the PV-to-BEV transformation implicitly.
One of the pioneering work employing this strategy is VED~\cite{lu2019monocular} which directly employ a variational auto-encoder to predict the BEV from an input image without intermediate stages.
To better preserve spatial information and to ease the integration of cross-view information, follow up works rely on more interpretable and elegant approaches to map the transformation between the features in the PV and the BEV.
One of the seminal work is~\cite{pan2020cross} where a Multi-layer Perceptron (MLP) is used to learn this mapping.
After the image features from multiple views are mapped onto a unified BEV, the segmentation can be learned into this final representation.
%
Unlike its IPM counterpart, it does not require any prior calibration and it is not affected by perspective distortion (global receptive field).
As a result, this strategy has influenced numerous works proposing various improvements such as multi-resolution features~\cite{roddick2020predicting, saha2021enabling} and combining geometric projection~\cite{li2021hdmapnet}.
More recently, to provide more expressive and data dependent mapping, the use of transformer~\cite{peng2022bevsegformer,yang2021projecting,li2022bevformer,chen2022efficient,zhou2022cross} has grown.

\subsubsection{Vectorized Map Construction}
The previously introduced literatures predict map elements in a rasterized BEV space.
The downside of this representation is its lack of structural relations and instance-level information.
In order to provide a lighter and more suitable representation for self-driving related downstream tasks~\cite{gao2020vectornet}, recent works~\cite{li2021hdmapnet,liu2022vectormapnet} propose estimating the vectorized HD map elements instead of a segmentation map, 
and InstaGraM belongs to this category.
A representative work is HDMapNet~\cite{li2021hdmapnet} which first predicts various representations as a rasterized BEV segmentation, then infers post-processing to generate vectorized representation of the map.
Despite promising results, the heuristic and handcrafted post-processing requires large amount of computations.
In order to predict a vectorized map in an end-to-end manner, VectorMapNet~\cite{liu2022vectormapnet} proposes two successive transformer decoders; the first decoder detects map elements via cross-attention between the BEV feature and element queries while the second transformer adopts auto-regressive decoder to recurrently generate polylines.
However, detection from element queries with cross-attention is known for its slow convergence, thus requires longer train epochs~\cite{carion2020end,zhu2020deformable}.
Auto-regressive decoder in polyline generator of VectorMapNet makes its computation heavy, which is not applicable for autonomous driving tasks that usually demand real-time computation.
In contrast, our proposed architecture does not require large amount of training time nor heavy computation of recurrent model.
MapTR, MapTRv2~\cite{liao2023maptr,liao2023maptrv2} further extend the transformer-based architecture with hierarchical query embedding and hierarchical set-to-set matching based on permutation-equivalent modeling of the map elements.
Although its higher efficiency with one-stage framework, the transformer-based architecture still suffers from scalability problem, whereas our proposed model with graph neural network maintains its high scalability as demonstrated in the experiments.
MapTRv2 enhances the prediction with faster convergence through auxiliary one-to-many matching and dense supervision.
Despite its strong baseline that encourages future research, we view this as a concurrent work with ours and show its results for reference.
We further demonstrate that our method achieves better performance against MapTRv2 under scalability experiments.

\subsection{Learning Structural Reconstruction}
Reconstructing structural data representation, such as polygon and polyline extraction, is one field this work falls into.
This field contains various sub-tasks such as line segment detection~\cite{zhou2019end,li2021ulsd} and polygon extraction of planar architectures~\cite{nauata2020vectorizing, zorzi2022polyworld, chen2022heat}.
L-CNN~\cite{zhou2019end} learns to detect lines and junctions of the scene using a single neural network.
It first detects junction heatmap from the CNN features of the input image and verifies the corresponding line proposals in an end-to-end manner.
ULSD~\cite{li2021ulsd} further extends the work to a unified line segment representation based on Bezier curve, enabling to model various shapes of line segment (i.e. distortion from camera model).
\cite{nauata2020vectorizing} proposes to reconstruct the planar graph of the outdoor building architecture from 2D satellite images.
The model first detects three geometric primitives, associates their relationships, and utilizes Integer Programming that optimizes the geometric primitives to finally reconstruct building polygons.
HEAT~\cite{chen2022heat} exploits both image and geometric information via attention-based neural network to reconstruct a planar graph of the sturctures.
Similar to prior works, it first detects corners by predicting confidence map and filters the corners by non-maximum suppression to select corner candidates.
The deformable attention mechanism of HEAT fuses geometric coordinates of the detected corners with image feature to classify edge candidates, which are initialized by pairing all detected corners.
Despite its compelling performance by holistic edge attention mechanism, it suffers from high computational cost by pairing all corner candidates, which is mostly larger than the underlying corners, and feeding to the memory intensive transformer.
As its counterpart, the proposed InstaGraM only takes the vertex (corner) and its corresponding node embedding to predict their connection (edge) via graph neural network, extending its scalability.
Furthermore, compared to aforementioned works, our proposed network can model both polylines (\textit{e.g.} lane divider) and polygons (\textit{e.g.} island-shaped road boundary) thanks to a unified instance-level graph modeling.

A work sharing similarities with our approach is PolyWorld~\cite{zorzi2022polyworld} which predicts the buildings' contours as a set of polygons from satellite images.
Similarly to our strategy, this technique uses a CNN for vertex detection followed by a GNN for association.
In contrast with PolyWorld, we adopt the interest point decoder from~\cite{detone2018superpoint} predicting high-resolution vertex coordinates.
Furthermore, we leverage the distance transform embedding for implicit directional information between vertices to associate.
Finally, our strategy is designed for road elements detection requiring both semantic and instance-level information.

%% file: Section/3_method.tex
\section{Method} 
\label{sec:method}
\begin{figure*}[tb]
\centering
\includegraphics[width=0.9\linewidth]{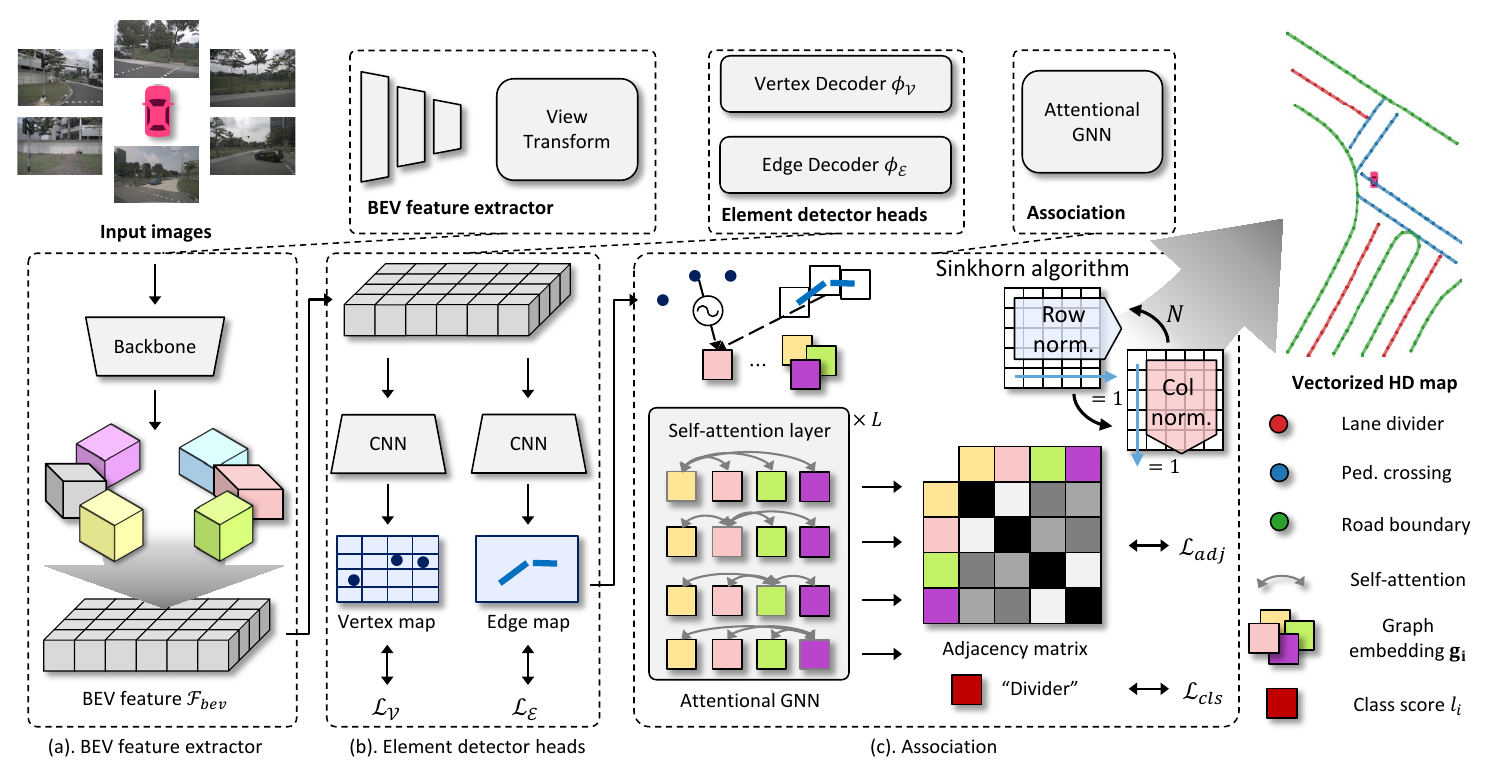}
\vspace{-5pt}
\caption{Proposed InstaGraM architecture. 
The blocks at the top show the overall components of InstaGraM architecture and the bottom blocks show the details of structure and training of each component.
}
\label{fig:3_Architecture}
\vspace{-12pt}
\end{figure*}

We propose an end-to-end network to compute a BEV vectorized HD map from a set of cameras mounted on a vehicle. 
To represent road elements (\textit{i.e.}, lane dividers, pedestrian crossing, and road boundaries), HD maps typically consist of 2D vectorized vertices and their instance-level adjacency connectivity.
To obtain this vector representation, previous works rely on segmentation prediction and heavy post-processing~\cite{li2021hdmapnet}, auto-regressive models~\cite{liu2022vectormapnet}, or direct set prediction using Transformer decoder~\cite{liao2023maptr} known for their high computational cost.
In contrast, we propose a lighter pipeline based on a combination of CNNs and a GNN able to predict a set of vertices and their adjacency directly.
Our method is three-folded.
First, similar to recent BEV framework, we utilize relevant view transform approach to build a unified BEV features from the CNN features extracted from each image captured from the camera rig -- via an EfficientNet~\cite{tan2019efficientnet}.
From this BEV feature map, two CNN decoders extract the vertices and edge maps of the observed road elements.
Finally, these vertices and their local edge response are fed to an attentional GNN in order to learn the semantic class and the connection between the vertices.

\subsection{View Transform}\label{method:transform}
The very first stage of our HD map estimation network is extraction of the top-down BEV features map~$\mathcal{F}_{bev}$ by combining the CNN features~\cite{tan2019efficientnet} from each images captured by the camera rig at a given time.
We adopt various off-the-shelf view transform methods to transform perspective view features to a unified BEV space.
We conduct experiments by choosing recent view transform and show that our work can be well adopted to various methods, which enhances scalability of the proposed architecture.
For more information regarding the extraction of this feature map, we refer to~\cite{philion2020lift,li2021hdmapnet} that we have carefully replicated for this stage.

\subsection{Element Detector Heads}\label{method:detector}
From the top-down feature map $\mathcal{F}_{bev}$, we extract the \emph{vertices} and \emph{edges} of the HD map elements using two CNN decoders $\phi_{\mathcal{V}},\phi_{\mathcal{E}}$ respectively.
These two components are predicted in the rasterized BEV space $\mathbb{R}^{W_{bev}\times H_{bev}}$ similar to segmentation tasks.
The vertex decoder $\phi_{\mathcal{V}}$ adopts the interest point decoder from~\cite{detone2018superpoint} and extracts possible position heatmap at every $S\times S$ local, non-overlapping grid in BEV pixels.
It computes $\mathcal{X}\in\mathbb{R}^{{W_{c}}\times {H_{c}}\times(S^2+1)}$, where $\{W_c\times H_c\}=\{\frac{W_{bev}}{S}\times \frac{H_{bev}}{S}\}$ denotes the number of grids and the $(S^2+1)$ channels indicating possible position in the local grids with an additional ``no vertex" dustbin.
After a channel-wise softmax, the dustbin dimension is removed and the vertex heatmap is reshaped from $\mathbb{R}^{W_c\times H_c\times(S^2)}$ to $\mathbb{R}^{W_{bev}\times H_{bev}}$.
We analyze the effect of the resolution of BEV ($\{W_{bev}\times H_{bev}\}$) and the local grid size $S$ in Section~\ref{sec:experiments}.
In parallel with the vertex decoder, the edge map decoder $\phi_{\mathcal{E}}$ predicts the distance transform map $\mathcal{D}\in\mathbb{R}^{W_{bev}\times H_{bev}\times C}$, the $C$ channels indicating the number of class categories of map elements.
The distance transform provides strong prior to geometry of map elements by assigning the distances of BEV grids to the nearest map element.
This edge map of distance transform~\cite{borgefors1986distance} implicitly provides spatial relations between vertices and curvature of map elements inspired by Signed Distance Function (SDF) from 3D applications~\cite{park2019deepsdf, liu2020dist, wang2021neus}.
We further demonstrate in Section~\ref{sec:experiments} that this distance transform representation as an edge map plays a significant role in instance-level association.
We apply ReLU and a threshold after the last Conv layer to predict the distance values from 0 to 10 in the rasterized BEV image.

\subsection{Association via Graph Neural Network}\label{method:graph}
The two components extracted from element detector heads are associated via a graph neural network, where all vertices interact throughout an attention scheme~\cite{vaswani2017attention,dosovitskiy2020image}.
This allows our network to reason about both point-level and instance-level relations between map elements based on various attributes including positions, implicit edge map of distance values and class categories.

\begin{figure}[t]
\centering
\includegraphics[width=0.9\linewidth]{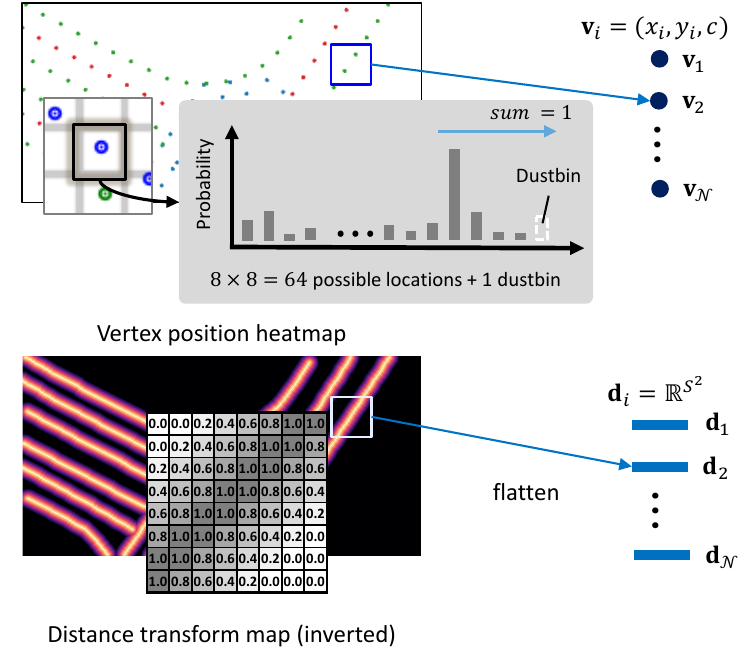}
\vspace{-5pt}
\caption{Illustration of graph embedding extraction.
Vertex position (top) provides geometric information of vertices, whereas distance transform map (bottom) supports local connectivity between vertices as well as the spatial structure of map elements via weights distributed along the normal direction of connection centered at map elements.
}
\label{fig:4_Distance}
\vspace{-12pt}
\end{figure}

\noindent\textbf{Graph Embeddings:}
We combine vertex positions and distance transform maps to form initial graph embeddings.
We first extract the position of each vertex in rasterized BEV coordinate and their respective confidence from channel-wise softmax in the vertex position heatmap, $\mathbf{v}_i=(x_i,y_i,c_i)$.
We only extract one distinctive vertex position with maximum confidence in each $S\times S$ grid cell, which is acting similar to Non-Maximum Suppression.
After extraction, a $i$-th vertex position $\mathbf{v}_i$ is encoded by a sinusodial positional encoding function $\gamma$ to augment it into a high-dimensional vector~\cite{mildenhall2021nerf}.
This positional encoding is further supported by an additional shallow MLP.
To complement the positional information of the vertex $\mathbf{v}_i$, we additionally include the local directional information $\mathbf{d}_i$ as the embedding of the distance transform patch corresponding to the same grid cell (hence $\mathbf{d}_i\in\mathbb{R}^{S^2}$).
The extraction of vertex and distance transform patch is illustrated in Fig.~\ref{fig:4_Distance}.
Then our initial graph consists of $D$-dimensional embeddings $^{(0)}\mathbf{g}_i\in\mathbb{R}^D$, combining both the vertex position and its directional local information, can be formulated as:

\begin{equation}
    \label{eq:embedding}
    ^{(0)}\mathbf{g}_i=\text{MLP}_{\mathcal{V}}(\gamma(\mathbf{v}_i))+\text{MLP}_{\mathcal{E}}(\mathbf{d}_i).
\end{equation}
This enables us to associate multiple graph embeddings based on their vertex and edge representation throughout an attention scheme afterward.

\noindent\textbf{Attentional Message Passing:}
We start from an initial graph $^{(0)}\mathcal{G}$ with nodes containing both vertex position and edge map embeddings as a high-dimensional vector.
This initial graph has bidirectional edges, connecting $i$-th vertex to all other vertices.
To further enhance the nodes and find the final edges of the vertices, we pass the initial graph to the attentional graph neural network and propagate this graph through message passing~\cite{velivckovic2017graph,sarlin2020superglue}.
Our objective is to find final bidirectional edges of the vertices as an instance-level information of map elements.
We feed our initial graph to attentional graph neural network that aggregates graph embeddings via a message passing consists of MLP and Multi-head Self Attention (MSA):
\begin{equation}
    \label{eq:gnn}
    \begin{aligned}
        ^{(0)}\mathcal{G}&=[^{(0)}\mathbf{g}_1;^{(0)}\mathbf{g}_2;\dots;^{(0)}\mathbf{g}_\mathcal{N}] \\
        ^{(l)}\mathcal{G}&={^{(l-1)}\mathcal{G}}+\text{MLP}({[{{^{(l-1)}\mathcal{G}}\Vert\text{MSA}(^{(l-1)}\mathcal{G})}]}),l=1,\dots,L
    \end{aligned}
\end{equation}

Self-attention and aggregation in Eq.~\ref{eq:gnn} provide interaction between all the graph embeddings based on their spatial and directional appearance embedded in $\mathbf{g}_i$.
Concretely, each vertex node attends to all other nodes to find the next possible vertices that would appear in the map.
After $L$ layers of attentional aggregation, class scores $\mathbf{l}_i\in\mathbb{R}^3$ and graph matching embeddings $\mathbf{f}_i\in\mathbb{R}^D$ are obtained:
\begin{equation}
    \label{eq:MLP}
    \begin{aligned}
        \mathbf{l}_i&=\text{MLP}_{cls}(^{(L)}\mathbf{g}_i) \\
        \mathbf{f}_i&=\text{MLP}_{match}(^{(L)}\mathbf{g}_i).
    \end{aligned}
\end{equation}

\noindent\textbf{Adjacency Matrix:}
We predict optimal edges by computing score matrix $\hat{\mathbf{S}}\in\mathbb{R}^{\mathcal{N}\times\mathcal{N}}$ between nodes of the graph $^{(L)}\mathcal{G}$.
The adjacency score between nodes $i$ and $j$ can be computed as cosine similarity of embedding vectors
\begin{equation}
    \label{eq:score}
    \hat{\mathbf{S}}_{ij}=<\mathbf{f}_i,\mathbf{f}_j>,\forall\{i,j\}\in\mathcal{N}\times\mathcal{N},
\end{equation}
where $<\cdot,\cdot>$ is an inner product of two embeddings.
Following SuperGlue~\cite{sarlin2020superglue}, we augment this score matrix to $\bar{\mathbf{S}}\in\mathbb{R}^{(\mathcal{N}+1)\times(\mathcal{N}+1)}$ with dustbin node for vertices that might not have any match, \textit{i.e.} a vertex at the end of an element instance.
The Sinkhorn algorithm~\cite{sinkhorn1967concerning,cuturi2013sinkhorn} that iteratively normalizes $\exp{(\bar{\mathbf{S}})}$ along rows and columns is used to compute final adjacency matrix of the graph.
Adjacency matrix $\bar{\mathbf{A}}\in\mathbb{R}^{(\mathcal{N}+1)\times(\mathcal{N}+1)}$ with instance-level edges can be computed throughout this optimal matching with augmented score $\bar{\mathbf{S}}$.
Once the optimal adjacency matrix $\bar{\mathbf{A}}$ is obtained, instance-level elements can be inferred by simply following adjacency flow of vertices until they meet the dustbin node.
\subsection{Losses}\label{method:losses}
We design the whole network to be differentiable, allowing us to train it in a fully supervised manner with combinations of losses at multiple branches.
For supervision of element detector heads, cross-entropy with softmax loss and L2 loss are used for vertex location heatmap and distance transform map respectively.
\begin{equation}
    \label{eq:cnn_loss}
    \begin{aligned}
        \mathcal{L}_{\mathcal{V}}(\mathcal{X},\mathcal{Y}) & = {\frac{1}{ H_c,W_c}}\sum_{h=1,w=1}^{H_c,W_c}l_p(\mathbf{x}_{hw};y_{hw}) \\
        \mathcal{L}_{\mathcal{E}}(\hat{\mathcal{D}},\mathcal{D}) & = {\frac{1}{N}}\sum_{d_p\in\mathcal{D}}\Vert d_p-\hat{d_p}\Vert^2,
    \end{aligned}
\end{equation}
where $H_c=\frac{H_{bev}}{S}$ and $W_c=\frac{W_{bev}}{S}$ are the indexing dimension of $S\times S$ local cells.

The coordinates from the vertex location heatmap prediction may not align perfectly to the ground truth vertex coordinates, specifically in the early stage of training, resulting in ambiguity of the ground truth adjacency and class label.
To address this we find the nearest pairs between the ground truth vertices and predicted vertices to provide the ground truth for the output of the graph neural network, adjacency matrix and class predictions.
The nearest ground truth vertex $\sigma(i)$ to the predicted vertex $i$ is obtained that minimizes the Chamfer distance cost with threshold $D_0$:
\begin{equation}
    \label{eq:gt_pair}
    \sigma=\argmin{D(\mathbf{v}_i,\mathbf{v}_{\sigma(i)})<D_{0}}\sum_{i}^{\mathcal{N}}D(\mathbf{v}_i,\mathbf{v}_{\sigma(i)}).
\end{equation}
From the matching pairs $[\mathbf{v}_i,\mathbf{v}_{\sigma(i)}]$, the ground truth adjacency pairs $\mathcal{M}=\{(i,j)\}\in\mathds{1}^{\mathcal{N}\times(\mathcal{N}+1)}$ between the vertices $i$ and $j$ are obtained by observing connection between the ground truth vertices $\sigma(i)$ and $\sigma(j)$.
The predicted vertices that do not have the ground truth pair within $\sigma$ fall into a dustbin vertex $\varnothing$ as ground truth adjacency.
Since the vectorized representation of the map has bidirectional edges as illustrated in~Fig.~\ref{fig:1_Graph}, we compute negative log-likelihood of adjacency loss for both forward and backward direction $\mathcal{M}=\{(i,j)\}\in\mathds{1}^{\mathcal{N}\times(\mathcal{N}+1)}$ and $\mathcal{M}^T=\{(j,i)\}\in\mathds{1}^{\mathcal{N}\times(\mathcal{N}+1)}$.
\begin{equation}
    \label{eq:adj_loss}
    \mathcal{L}_{adj}=-\frac{1}{2}    \left(\sum_{(i,j)\in\mathcal{M}}\log\Bar{\mathbf{A}}_{ij}+\sum_{(i,j)\in\mathcal{M}^T}\log\Bar{\mathbf{A}}_{ij}\right).
\end{equation}
We further supervise the graph neural network with negative log-likelihood for vertex classification.
Through this supervision, our graph neural network can reason about the vertex label categories in addition to the instance-level information.
\begin{equation}
    \label{eq:gnn_loss}
    \begin{aligned}
        \mathcal{L}_{cls}&=\sum_{i}^{\mathcal{N}}\log{l_{\sigma(i)}}. \\
    \end{aligned}
\end{equation}
Combining above losses, our final loss function is formulated as:
\begin{equation}
    \label{eq:loss}
    \mathcal{L}=\lambda_1\mathcal{L}_{\mathcal{V}}+\lambda_2\mathcal{L}_{\mathcal{E}}+\lambda_3\mathcal{L}_{adj}+\lambda_4\mathcal{L}_{cls},
\end{equation}
where we set $\lambda_3,\lambda_4,\ll\lambda_1,\lambda_2$ to ensure the graph neural network have enough vertex predictions to associate, specifically during the early stage of training.

%% file: Section/4_exp.tex
\section{Experiments}
\label{sec:experiments}

\subsection{Dataset and Evaluation}
We train and evaluate our network on the nuScenes dataset~\cite{holger2020nuscenes}, which consists of 1000 scenes of 20 seconds captured from a vehicle equipped with cameras, LIDARs, IMU, and GPS.
Aside from sensory data, this dataset also has the specificity to provide fully annotated HD maps.
From each sequence in the dataset, keyframes containing the pose in the global HD map, 3D point clouds, and images are extracted at 2Hz.
For our application, we solely use the images captured by the six surrounding cameras mounted on the vehicle and the HD map.
In order to train our pipeline with the local HD map expressed in the vehicle's referential, we use the data processing scheme proposed in~\cite{li2021hdmapnet}.
From this data, three kinds of road elements are kept in the sampled local HD maps: lane dividers, pedestrian crossings, and road boundaries.
Additionally, to train the intermediate representation of our network, we generate the ground truth vertex map and distance transform map from the local HD map.
Regarding the ground-truth vertices generation, we follow~\cite{detone2018superpoint} by subsampling one vertex for each non-overlapping $8\times8$ grid region in the BEV grids.
We use the OpenCV library to generate the distance transform map within a distance range $[0,10]$.
For a fair comparison with the baselines, we adopt the Average Precision (AP) of instance prediction as the evaluation metric with the Chamfer distance thresholds $\{0.5, 1.0, 1.5\}$.

\subsection{Implementation Details}
Our model is trained and evaluated using PyTorch, on RTX 3090 GPUs with batch size $32$.
We define the perception range $[-15.0m,15.0m]$ along the $X$-axis and $[-30.0m,30.0m]$ along the $Y$-axis in the vehicle's referential.
We rasterize this range with resolution $0.15m$ resulting in $H_{bev}=200$ and $W_{bev}=400$ as output of the element detector heads.
We adopt EfficientNet~\cite{tan2019efficientnet} as backbone.
Neural view transform maps 6 image features to a BEV feature map $\mathcal{F}_{bev}\in\mathbb{R}^{{\frac{W_{bev}}{2} \times \frac{H_{bev}}{2}\times256}}$.
The element detector head utilizes 3 Residual blocks~\cite{he2016deep} and upsampling to predict the vertex location heatmap and the distance transform map.
We extract the top $\mathcal{N}=400$ vertices of confidences higher than threshold $0.01$ from softmax-ed vertex location heatmap.
The local distance transform patch of $S\times S$ grid is extracted that corresponds to vertex $\mathbf{v}_i$.
For each detected vertex, their respective local distance transform patch of size $S\times S$ is extracted to build the graph embeddings.
If the number of extracted vertices is inferior to $\mathcal{N}$, we pad the embeddings with zeros and adopt a mask operation within the graph neural network and the Sinkhorn iterations.
We force the diagonal entities of the score matrix $\bar{\mathbf{S}}$ to zeros before it is fed into the Sinkhorn algorithm -- since the map elements do not have ``self-loop".
For training, we set the loss scales $\lambda_1,\lambda_2,\lambda_3$ and $\lambda_4$ as $1.0,1.0,5e^{-3}$ and $1e^{-2}$ respectively.
During inference, we follow the flow of vertex connection in adjacency matrix to produce final vectorized format of the map elements.

\begin{table*}[tb]
\begin{minipage}{\textwidth}
    \centering
    \caption[]{%
    Evaluation results on nuScenes dataset.
    $\dagger$ denotes the results are reproduced using the public open code, adopting the same vectorized format.
    }
    \vspace{-5pt}
    \label{tab:1_Comparison}
    \resizebox{0.95\textwidth}{!}
    {
    \setlength{\tabcolsep}{6pt}
    \renewcommand{\arraystretch}{1.3}
    \begin{tabular}{l|ccc|cccc|c}
        \toprule[1pt]
        Method & Image size & Backbone\footnote{Effi and R denote EfficientNet and ResNet, respectively.} & Epochs & AP$_{\textit{div.}}$ & AP$_{\textit{ped.}}$ & AP$_{\textit{bound.}}$ & mAP & FPS\footnote{FPS of HDMapNet and VectorMapNet are provided by authors. We measure FPS of MapTR and InstaGraM on RTX 3090 GPU.
        } \\
        \hline
        \hline
        HDMapNet & $128\times352$ & Effi-B0 & 30 & 21.7 & 14.4 & 33.0 & 23.0 & 0.6 \\
        MapTR$^\dagger$ & $180\times320$ & R18 & 30 & 38.0 & \textbf{35.2} & \textbf{39.8} & \textbf{37.7} & \textbf{26.1} \\
        InstaGraM (Ours) & $128\times352$ & Effi-B0 & 30 & \textbf{41.5} & 30.2 & 35.9 & 35.9 & 24.8 \\
        \midrule
        VectorMapNet\footnote{Results with single stage training are used for fair comparison.} & $128\times352$ & R50 & 110 & 47.3 & 36.1 & 39.3 & 40.9 & 3.0 \\
        MapTR$^\dagger$ & $450\times800$ & R50 & 30 & 46.3 & 41.8 & \textbf{50.3} & 46.2 & 14.0 \\
        InstaGraM (Ours) & $128\times352$ & Effi-B4 & 30 & 48.3 & 33.9 & 43.5 & 41.9 & \textbf{19.2} \\
        InstaGraM (Ours) & $256\times704$ & Effi-B4 & 30 & \textbf{55.0} & \textbf{42.9} & 45.6 & \textbf{47.8} & 11.8 \\
        \midrule
        \color{gray}
        MapTRv2\color{black}$^\dagger$ & \color{gray}$450\times800$ & \color{gray}R50 & \color{gray}30 & \color{gray}63.2 & \color{gray}61.0 & \color{gray}63.3 & \color{gray}62.5 & \color{gray}12.0 \\
        \bottomrule[1pt]
    \end{tabular}
    }
    \vspace{-7pt}
\end{minipage}
\vspace{-12pt}
\end{table*}

\begin{table*}[tb]
    \centering
    \caption{Experiments on scalability.
    The perception range and input image size are reported in order of $X\times Y$ and $H\times W$, respectively.
    $\dagger$ denotes the results are reproduced adopting the same vectorized format and network configuration for fair comparison.
    }
    \vspace{-5pt}
    \label{tab:2_Scalability}
    \resizebox{0.95\textwidth}{!}
    {
    \setlength{\tabcolsep}{8pt}
    \renewcommand{\arraystretch}{1.3}
    \begin{tabular}{l|ccc|cc|c}
        \toprule[1pt]
        Method & Range & Image size & Backbone & AP$_{\textit{div.}}$ 
        & AP$_{\textit{bound.}}$ & mAP $_{\textit{div.}+\textit{bound.}}$ 
        \\
        \hline
        \hline
        MapTR$^\dagger$ & $60\times 30~m$ & $256\times704$ & Effi-B4 & 42.6
        & 45.4 & 44.0
        \\
        MapTRv2$^\dagger$ & $60\times 30~m$ & $256\times704$ & Effi-B4 & 50.7
        & \textbf{55.6} & \textbf{53.1}
        \\
        InstaGraM (Ours) & $60\times 30~m$ & $256\times704$ & Effi-B4 & \textbf{55.0} 
        & 45.6 & 50.3
        \\
        \midrule
        MapTR$^\dagger$ & $100\times100~m$ & $256\times704$ & Effi-B4 & 18.2 
        & 11.0 & 14.6
        \\
        MapTRv2$^\dagger$ & $100\times100~m$ & $256\times704$ & Effi-B4 & 25.1 
        & 16.9 & 21.0
        \\
        InstaGraM (Ours) & $100\times100~m$ & $256\times704$ & Effi-B4 & \textbf{34.4} 
        & \textbf{17.3} & \textbf{25.8}
        \\
        \bottomrule[1pt]
    \end{tabular}
    }
\vspace{-12pt}
\end{table*}

\begin{table}[t]
    \centering
    \caption{Ablation studies of the graph embeddings.
    }
    \vspace{-5pt}
    \label{tab:3_Ablation_embd}
    \resizebox{\columnwidth}{!}{
    \setlength{\tabcolsep}{3.5pt}
    \renewcommand{\arraystretch}{1.3}
    \begin{tabular}{c|cc|ccc|c}
        \toprule[1pt]
        InstaGraM & DT & PE & AP$_{\textit{divider}}$ & AP$_{\textit{ped}}$ & AP$_{\textit{boundary}}$ & mAP  \\ \hline\hline
        A      & \cmark & \xmark & 8.6 & 8.5 & 3.8 & 7.0 \\
        B      & \xmark & \cmark & 30.0 & 10.2 & 22.5 & 20.9 \\
        C      & \cmark & \cmark & 48.3 & 33.9 & 43.5 & 41.9 \\ 
        \bottomrule[1pt]
    \end{tabular}
    }
\vspace{-12pt}
\end{table}

\begin{table}[t]
    \centering
    \caption{Ablation studies on the number of GNN layers.
    }
    \vspace{-5pt}
    \label{tab:4_Ablation_gnn}
    \resizebox{\columnwidth}{!}{
    \setlength{\tabcolsep}{4.0pt}
    \renewcommand{\arraystretch}{1.3}
    \begin{tabular}{c|ccc|cc}
        \toprule[1pt]
        \# layers & AP$_{\textit{divider}}$ & AP$_{\textit{ped}}$ & AP$_{\textit{boundary}}$ & mAP & Param. \\ \hline\hline
        0 & 44.4 & 29.3 & 36.6 & 36.8 & 90.9M \\
        3 & 45.5 & 32.7 & 43.7 & 40.6 & 92.9M \\
        5 & 45.9 & 34.0 & 42.3 & 40.7 & 94.2M \\
        7 & {47.2} & {33.8} & {44.0} & {41.7} & 95.5M \\
        9 & 47.6 & 33.4 & {44.8} & 41.9 & 96.8M \\
        11 & {47.7} & {34.0} & 44.2 & {42.0} & 98.2M \\ 
        \bottomrule[1pt]
    \end{tabular}
    }
\vspace{-12pt}
\end{table}

\begin{table}[t]
    \caption{Ablation study on View Transform methods.}
    \vspace{-5pt}
    \label{tab:5_BEV}
    \resizebox{\columnwidth}{!}{
    \setlength{\tabcolsep}{9.0pt}
    \renewcommand{\arraystretch}{1.3}
    \begin{tabular}{l|ccc}
        \toprule[1pt]
        {Method} & {mAP} & {FPS} & {Param.} \\
        \hline\hline
        Warping (IPM)            & 6.1 & 15.9 & 30.7M \\
        Depth (LSS~\cite{philion2020lift})          & 41.9 & 19.2 & 30.7M \\
        MLP (HDMapNet~\cite{li2021hdmapnet})        & 41.7 & 20.7 & 95.5M \\
        \bottomrule[1pt]
    \end{tabular}
    }
\vspace{-12pt}
\end{table}

\begin{table}[t]
    \centering
    \caption{
    Ablation studies on spatial configure of vertex heatmap.
    }
    \vspace{-5pt}
    \resizebox{\columnwidth}{!}{
    \setlength{\tabcolsep}{5.5pt}
    \renewcommand{\arraystretch}{1.3}
    \begin{tabular}{cc|ccc|c}
        \toprule[1pt]
         $Res.$ & $S$ & AP$_{\textit{divider}}$ & AP$_{\textit{ped}}$ & AP$_{\textit{boundary}}$ & mAP \\
         \hline\hline
         0.3  & 4  & 47.4 & 31.3 & 30.0 & 36.2 \\
         0.3  & 10 & 35.2 & 11.7 & 41.8 & 29.5 \\
         0.15 & 8  & 48.3 & 33.9 & 43.5 & 41.9 \\
         0.15 & 10 & 45.8 & 29.6 & 44.6 & 40.6 \\
        \bottomrule[1pt]
    \end{tabular}
    }
    \label{tab:6_Ablation_vertex}
\vspace{-12pt}
\end{table}

\begin{figure*}[tb]
\centering
\includegraphics[width=1.0\linewidth]{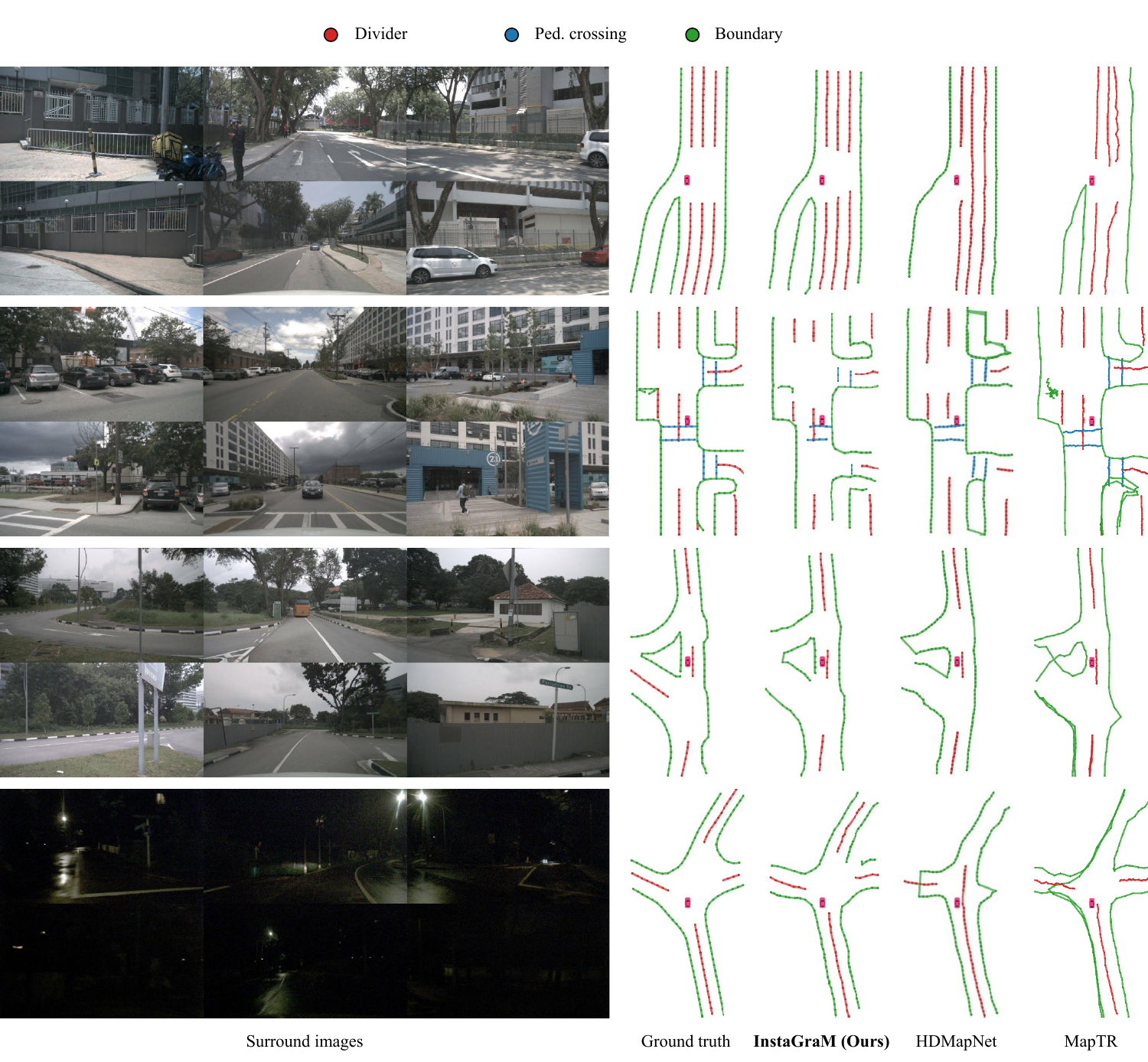}
\vspace{-5pt}
\caption{Qualitative comparison on complex traffic scenes under various conditions.
Beginning from the top row, sun, partly-cloud, cloud and night conditions.}
\label{fig:5_Results}
\vspace{-12pt}
\end{figure*}

\subsection{Results}
\noindent\textbf{Comparison with Baseline Methods.}
We compare our methods against previous vectorized HD map learning pipelines -- HDMapNet~\cite{li2021hdmapnet}, VectorMapNet~\cite{liu2022vectormapnet}, and MapTR~\cite{liao2023maptr}.
HDMapNet predicts 3 types of segmentation map --- semantic segmentation, instance embedding, and direction prediction --- and utilizes heuristic post-processing to generate vectorized map elements.
VectorMapNet utilizes transformer decoder from DETR~\cite{carion2020end} to predict the keypoints of the map elements, and autoregressive transformer decoder to generate detailed geometrical shape of the map elements.
MapTR adds heirarchical query embedding and permutation-equivalent set-to-set matching on top of DETR decoder.
We train and evaluate InstaGraM with EfficientNet-B0 and EfficientNet-B4, which are comparable backbones with HDMapNet and VectorMapNet/MapTR (ResNet-50) respectively.
\tableautorefname~\ref{tab:1_Comparison} shows that InstaGraM acheives $12.9$ higher mAP with $41\times$ faster inference speed compared to HDMapNet, and $1.0$ higher mAP with $6\times$ faster inference speed compared to VectorMapNet, under the comparable image size.
Furthermore, specifically in comparison with VectorMapNet, InstaGraM achieves better performance with significantly faster convergence time during training (train epochs of 110 for VectorMapNet and 30 for InstaGraM).
Compared to MapTR, InstaGraM achieves better performance with the comparable image size.
We add results of MapTRv2, state-of-the-art method in online vectorized HD map construction for reference, and we further tackle the scalability problem of MapTR and MapTRv2 in terms of accuracy against InstaGraM in \tableautorefname~\ref{tab:2_Scalability}.

\noindent\textbf{Scalability.}
We compare with the state-of-the-art vectorized HD map construction model, MapTR and MapTRv2~\cite{liao2023maptr, liao2023maptrv2}, at scalability configuration in \tableautorefname~\ref{tab:2_Scalability}, where we examine the accuracy of map construction at longer construction range.
We report results on long range setting, with $100m$ range along both X- and Y-axis, which is the typical range used for autonomous driving.
For the long range setting, we change the resolution of BEV feature map from $0.15$ to $0.25$, leading to feature size $200\times200$ to maintain the balance between computation and performance.
We increase the number of vertices $\mathcal{N}$ to $600$ for InstaGraM as the range increases, whereas for MapTR family, we increase the fixed number of vertices from 20 to 40 to handle the long range construction.
For MapTRv2, we use its full architectures that show the highest performance and only modify the range, image size and backbone.
\tableautorefname~\ref{tab:2_Scalability} shows that InstaGraM outperforms MapTR and state-of-the-art MapTRv2 by 11.2 mAP and 4.8 mAP respectively for \textit{divider} and \textit{boundary} classes, which appear over large area in long-range setting.
The querying method by the Transformer decoder inherently limits its scalability and due to the fixed number of query in design, it is hard to maintain the complex shape of various map elements, as shown in Fig.~\ref{fig:5_Results}.

\begin{figure*}[tb]
\centering
\includegraphics[width=1.0\linewidth]{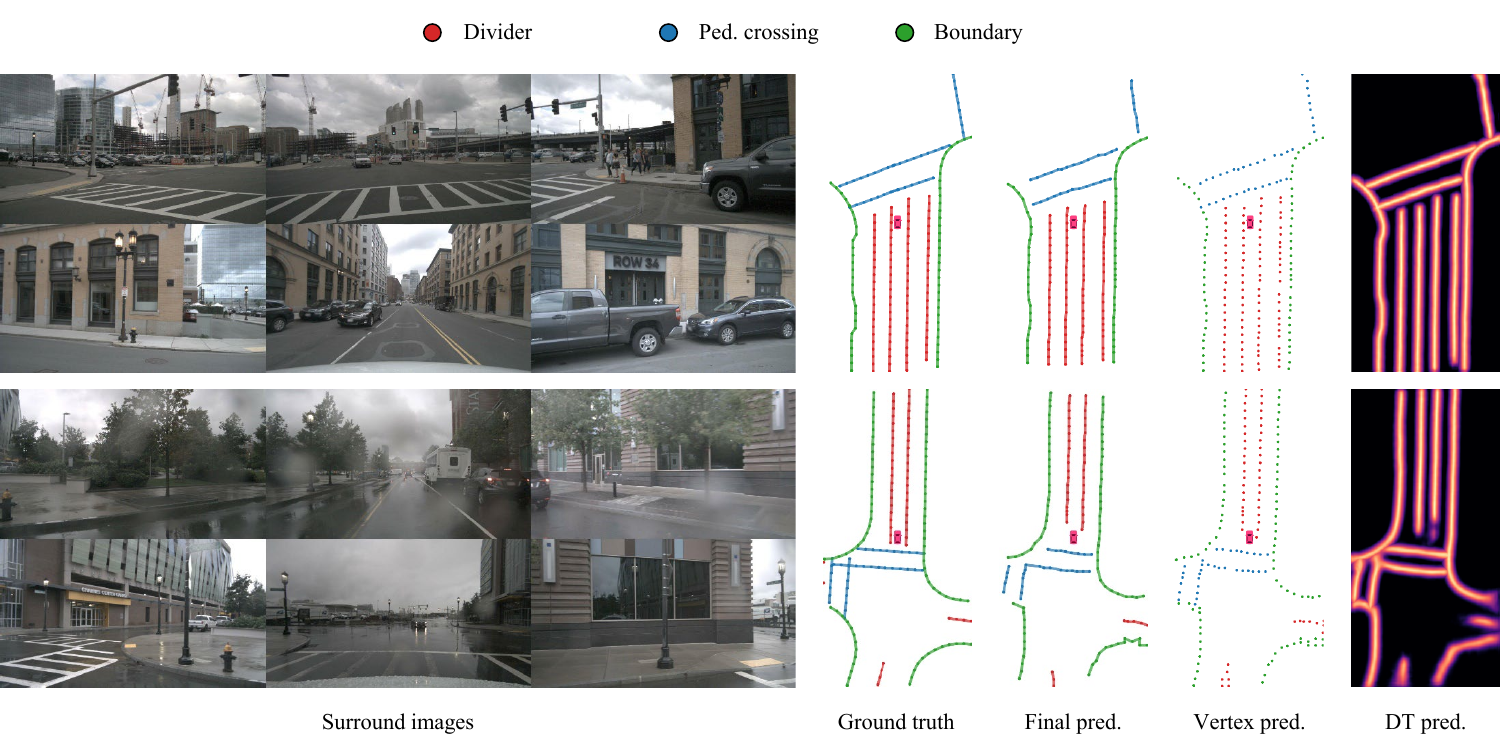}
\vspace{-5pt}
\caption{Visualization of intermediate results of InstaGraM on complex intersections.
Vertex and DT predictions are associated in the attentional GNN to produce the final prediction. Vertices in vertex prediction is colorized with classification result after the attentional GNN.}
\label{fig:6_Intermediate}
\vspace{-12pt}
\end{figure*}

\begin{figure}[tb]
\centering
\includegraphics[width=1.0\linewidth]{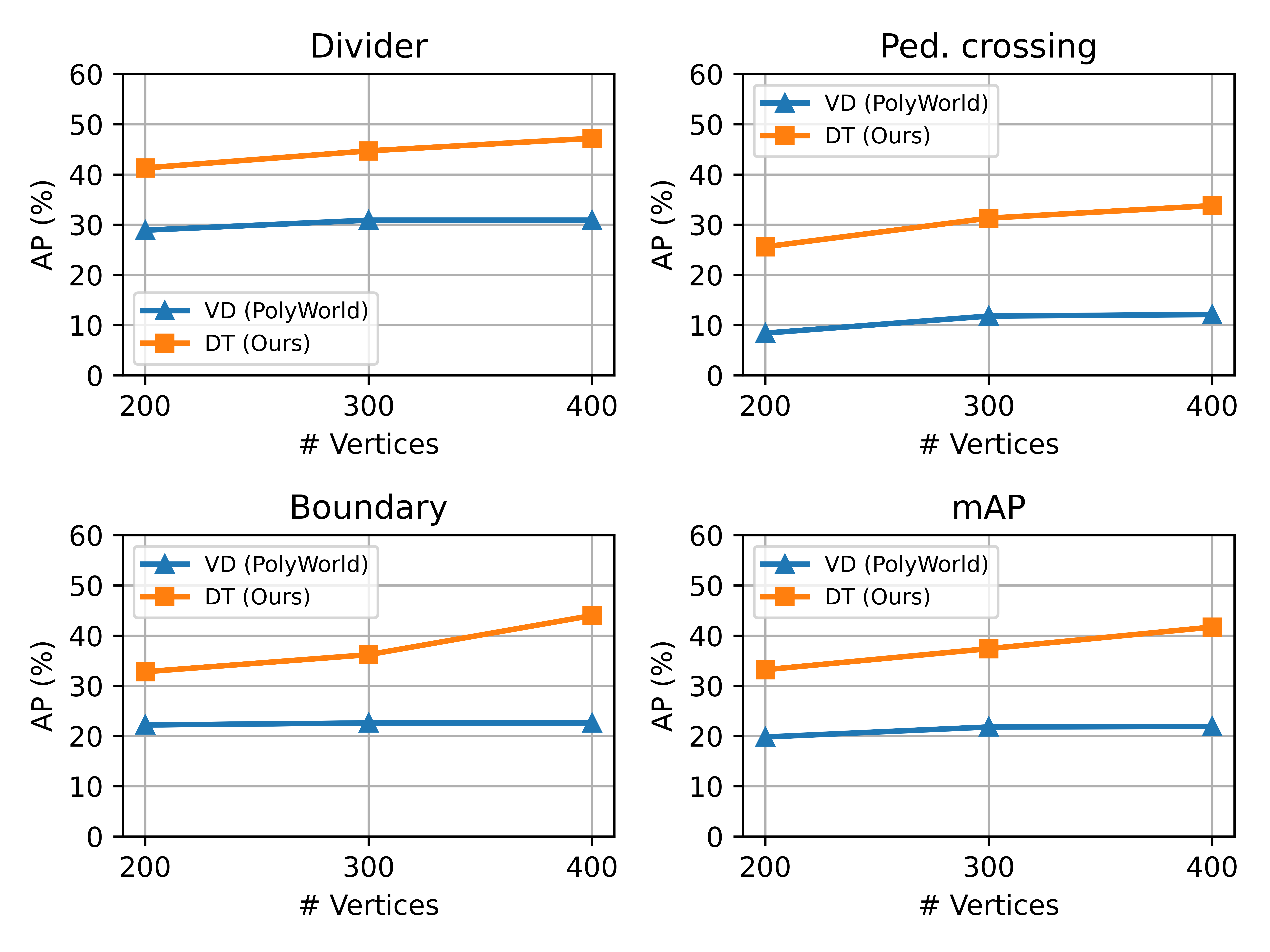}
\vspace{-5pt}
\caption{We compare InstaGraM with two different embeddings: the distance transform embeddings and the visual descriptor embeddings extracted from feature map.
VD denotes that the visual descriptor is used for graph embeddings similar to PolyWorld~\cite{zorzi2022polyworld}.
We compare two embeddings over various number of vertices extracted from vertex location heatmap.}
\label{fig:7_Ablation}
\vspace{-12pt}
\end{figure}

\noindent\textbf{Qualitative Results.}
We visualize the vectorized HD map predictions in Fig.~\ref{fig:5_Results}.
It demonstrates that InstaGraM generalizes well under various weather conditions.
InstaGraM computes, in an end-to-end manner without post-processing nor heavy computation, the semantic and instance-level information of the complex map elements.
Our graph modeling strategy shown in Fig.~\ref{fig:1_Graph} enables accurate and fast prediction of the map elements.
The distance transform and positional embedding provides precise primitives to associate through the graph neural network, able to predict various curve shapes, as additionally demonstrated in Fig.~\ref{fig:6_Intermediate}.
Moreover, the proposed InstaGraM can preserve complex shapes of the map elements, while HDMapNet and MapTR fail.

\noindent\textbf{Graph Embeddings.}
We conduct ablation studies for design choices of the graph embeddings as shown in \tableautorefname~\ref{tab:3_Ablation_embd}.
We find that from A and C in \tableautorefname~\ref{tab:3_Ablation_embd}, positional embedding plays significant role since the distance transform distribution appear the same in overall vertices (see Fig.~\ref{fig:2_Pipeline}).
It provides the graph neural network where to look for nearby vertices to associate.
The distance transform embedding further improves by providing strong directional information of the vertices (see B and C in \tableautorefname~\ref{tab:3_Ablation_embd}).

\noindent\textbf{Ablation Studies.}
We further conduct ablation studies with various number of graph neural network layers and different BEV transformation methods.
\tableautorefname~\ref{tab:4_Ablation_gnn} analyzes affects of the number of attention layers in our graph neural network.
Increasing the number of attention layers improves the accuracy of our model with saturation at 7 layers.
Proposed InstaGraM can be adapted several BEV transformation methods as shown in \tableautorefname~\ref{tab:5_BEV}.
We analyze the performance varying the resolution of rasterized BEV space and the local grid size $S$ for the position heatmap of the vertices.
We vary the resolution of the BEV space to determine the number of BEV pixels within the same range, $i.e.$ $\{W_{bev}\times H_{bev}\}=\{200\times 100\}$ for $Res.=0.3$ and $\{W_{bev}\times H_{bev}\}=\{400\times 200\}$ for $Res.=0.15$.
The grid size $S$ determines the size of channel in channel-wise softmax to locate one distinct vertex in the local grid.
We choose $Res.=0.15$ and $S=8$ for its performance and computation (\tableautorefname~\ref{tab:6_Ablation_vertex}).

\noindent\textbf{Distance Transform.}
We further analyze the distance transform embedding by comparing with visual descriptor embedding similar to the embeddings used in PolyWorld~\cite{zorzi2022polyworld}.
PolyWorld extracts a vector from image feature map corresponding to the predicted vertex coordinate, and refer to it as ``visual descriptor".
Since the spatial size of the visual BEV feature map does not match the spatial size of the vertex heatmap passed by the interest point decoder, we interpolate the feature map to align the spatial size.
We take the visual descriptor from the interpolated feature map corresponding to the vertex coordinate, then encode to form the graph embedding.
Fig.~\ref{fig:7_Ablation} compares the performance of InstaGraM with different graph embeddings — one with the distance transform embeddings (DT) and one with the visual descriptor embeddings (VD) — over various number $\mathcal{N}$ of vertices extraction.
Positional embeddings are adopted in both setups.
Although visual descriptor from direct feature map contains high-level semantic information of the scene, the distance transform provides strong prior of the map elements, resulting in better performance.
We further emphasize that by adding auxiliary task of the distance transform regression, it implicitly provides the network the geometry of the ground truth map, outperforming baselines with fast convergence.

%% file: Section/5_conc.tex
\section{Conclusion}
\label{sec:conclusion}
We propose InstaGraM, an end-to-end vectorized HD map learning pipeline that is applicable for real-time autonomous driving.
We model the vectorized lines of HD map elements as a graph with vertices and adjacency matrix that composes instance-level edges.
Compared to previous works that require heuristic post-processing or a large amount of computational cost, InstaGraM computes in real-time polylines of the map elements achieving better performance compared to the state-of-the-art method.
We further demonstrate that our graph modeling and the GNN design has potential improvement for scalable autonomous driving.
This work focuses on HD map consturction at single frame.
To achieve complete HD map construction, the proposed work can be extended to video input stream, which we place as future work.

%% file: Section/6_bio.tex
 


\vspace{-15pt}
\begin{IEEEbiography}[{\includegraphics[width=1in,height=1.25in,clip,keepaspectratio]{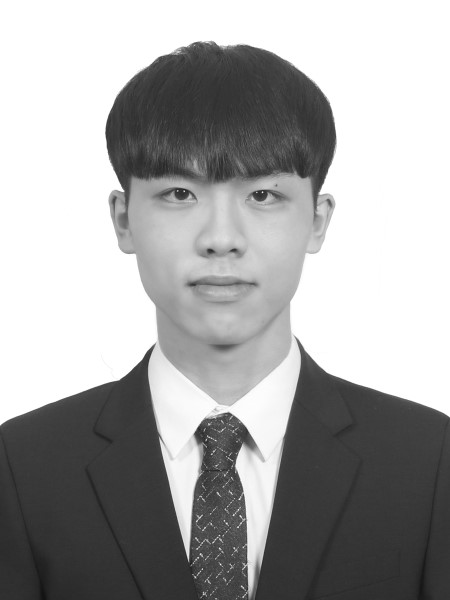}}]{Juyeb Shin}
received his B.S. degree in School of Robotics from Kwangwoon University, Seoul, South Korea, in 2021, and an M.S. degree from The Robotics Program, Korea Advanced Institute of Science and Technology (KAIST), Daejeon, South Korea, in 2023. He is currently pursuing a Ph.D. in The Robotics Program, KAIST. His research interests include computer vision and deep learning, especially focusing on mapping and localization for autonomous vehicles.
\end{IEEEbiography}
\vspace{-15pt}

\begin{IEEEbiography}[{\includegraphics[width=1in,height=1.25in,clip,keepaspectratio]{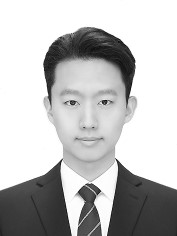}}]{Hyeonjun Jeong}
received his B.S. degree in the Department of Automotive Engineering from Kookmin University, Seoul, South Korea, in 2021, and an M.S. degree from the Graduate School of Mobility, Korea Advanced Institute of Science and Technology (KAIST), Daejeon, South Korea, in 2023. He is currently pursuing a Ph.D. in the Graduate School of Mobility, KAIST. His research interests include computer vision and deep learning, especially focusing on self-supervised learning for autonomous vehicles.
\end{IEEEbiography}
\vspace{-15pt}

\begin{IEEEbiography}[{\includegraphics[width=1in,height=1.25in,clip,keepaspectratio]{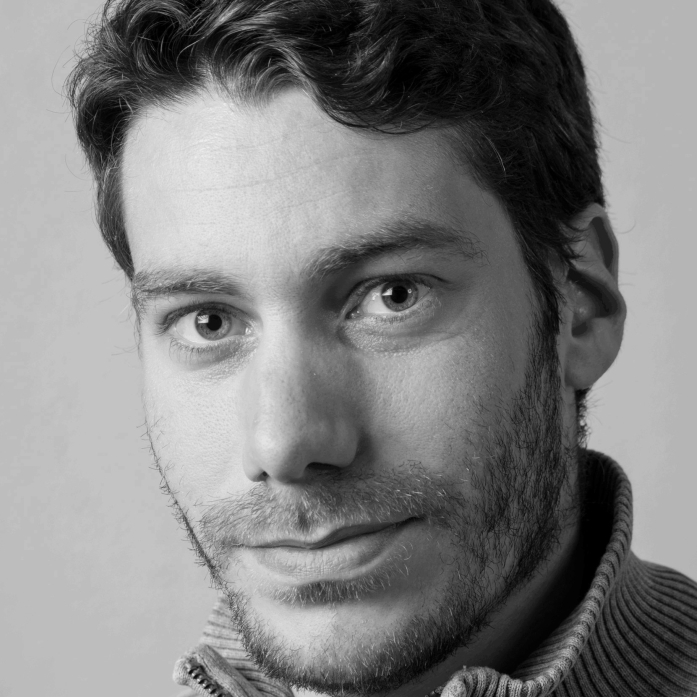}}]{Francois Rameau} is an Assistant Professor at the State University of New York (SUNY) in Korea.
He received his Ph.D. in Vision and Robotics from the University of Burgundy (France) in 2014 under the joined supervision of Prof. Cedric Demonceaux, Prof. Desire Sidibe, and Prof. David Fofi. Later. He joined the Korea Advanced Institute of Science and Technology (KAIST, South Korea), first as a postdoctoral researcher and then as a Research Professor under the KRF fellowship program (2017-2023). 
Dr. Rameau's research interests center around 3D Computer Vision, Machine Learning, and Collaborative Robotics. He has an accomplished research track record, with numerous publications in top-tier venues.
\end{IEEEbiography}
\vspace{-15pt}

\begin{IEEEbiography}[{\includegraphics[width=1in,height=1.25in,clip,keepaspectratio]{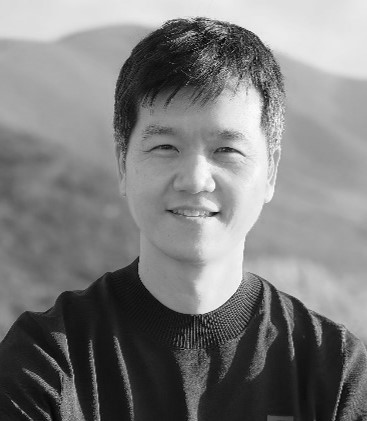}}]{Dongsuk Kum}
received his Ph.D. degree in mechanical engineering from the University of Michigan, Ann Arbor, MI, USA, in 2010. He was a Visiting Research Scientist with the General Motors Research and Development Propulsion Systems Research Laboratory, Warren, MI, USA, where he focused on advanced propulsion system technologies, including hybrid electric vehicles, flywheel hybrids, and waste heat recovery systems. He is currently an Associate Professor with the Graduate School of Mobility, Korea Advanced Institute of Science and Technology, where he is also the Director of the Vehicular Systems Design and Control Laboratory. His research centers on the modeling, control, and design of advanced vehicular systems with particular interest in hybrid electric vehicles and autonomous vehicles.
\end{IEEEbiography}
